\documentclass{article}

\PassOptionsToPackage{numbers,compress}{natbib}
\usepackage[preprint]{neurips_2025}

\usepackage[utf8]{inputenc} 
\usepackage[T1]{fontenc}    
\usepackage{hyperref}       
\usepackage{url}            
\usepackage{booktabs}       
\usepackage{amsfonts}       
\usepackage{nicefrac}       
\usepackage{microtype}   
\usepackage{xcolor}         
\usepackage{amsmath}
\usepackage{float}
\usepackage{graphicx} 
\usepackage{booktabs}
\usepackage{multirow} 
\usepackage{makecell}
\usepackage{subcaption}
\usepackage{caption}                 
\title{Distill CLIP (DCLIP): Enhancing Image-Text Retrieval via Cross-Modal Transformer Distillation}

\author{%
Daniel Csizmadia\textsuperscript{1}\thanks{First Author}\quad
Andrei Codreanu\textsuperscript{1}\quad
Victor Sim\textsuperscript{1}\quad
Vighnesh Prabhu\textsuperscript{1}\\
\textbf{Michael Lu}\textsuperscript{1}\thanks{Senior Author}\quad
\textbf{Kevin Zhu}\textsuperscript{1}\footnotemark[2]\quad
\textbf{Sean O'Brien}\textsuperscript{1}\footnotemark[2]\quad
\textbf{Vasu Sharma}\textsuperscript{1,2}\footnotemark[2]\\
\textsuperscript{1}Algoverse AI Research\quad
\textsuperscript{2}Carnegie Mellon University\\
\texttt{csizmadiadaniel@vt.edu, kevin@algoverse.us}
}

\begin{document}

\maketitle

\begin{abstract}

We present \textbf{Distill CLIP (DCLIP)}, a fine-tuned variant of the CLIP model that enhances multimodal image-text retrieval while preserving the original model's strong zero-shot classification capabilities. CLIP models are typically constrained by fixed image resolutions and limited context, which can hinder their effectiveness in retrieval tasks that require fine-grained cross-modal understanding. DCLIP addresses these challenges through a meta teacher-student distillation framework, where a cross-modal transformer teacher is fine-tuned to produce enriched embeddings via bidirectional cross-attention between YOLO-extracted image regions and corresponding textual spans. These semantically and spatially aligned global representations guide the training of a lightweight student model using a hybrid loss that combines contrastive learning and cosine similarity objectives. Despite being trained on only $\sim$67{,}500 samples curated from MSCOCO, Flickr30k, and Conceptual Captions—just a fraction of CLIP’s original dataset—DCLIP significantly improves image-text retrieval metrics (Recall@K, MAP), while retaining approximately 94\% of CLIP’s zero-shot classification performance. These results demonstrate that DCLIP effectively mitigates the trade-off between task specialization and generalization, offering a resource-efficient, domain-adaptive, and detail-sensitive solution for advanced vision-language tasks. Code available at \url{https://anonymous.4open.science/r/DCLIP-B772/README.md}.

\end{abstract}

\section{Introduction}

Contrastive vision-language models such as CLIP~\cite{clip} have demonstrated strong performance in zero-shot classification and image-text retrieval by leveraging large-scale noisy image-caption pairs. However, their reliance on global alignment between full images and texts limits fine-grained understanding—making it difficult to localize object attributes or model complex visual relationships critical for open-domain retrieval and compositional reasoning.

Recent works have addressed this by incorporating region-level alignment through dense grounding, prompt engineering, or architectural modifications. While effective, these often require heavy supervision (e.g., bounding boxes) or sacrifice CLIP's simplicity and efficiency. Models like LongCLIP~\cite{longclip} and TULIP~\cite{tulip} relax CLIP’s input limitations, but efficient fine-grained retrieval with strong generalization remains a challenge.

We propose Distill CLIP (DCLIP), a lightweight framework for improving fine-grained retrieval while preserving CLIP’s zero-shot capabilities. Instead of full retraining or complex fusion architectures, DCLIP uses a targeted distillation setup: a meta-teacher enriches CLIP’s image embeddings via region-level attention, and a student model partially fine-tunes its image encoder to mimic these outputs. The text encoder remains frozen, maintaining alignment with CLIP’s original semantic space.

DCLIP extracts coarse regions via YOLO and applies bidirectional cross-modal attention to generate fine-grained embeddings. These serve as distillation targets for the student, which is trained using contrastive, cosine, and anchor-based losses. This design enables improved representation learning without requiring region processing at inference or risking generalization collapse from aggressive fine-tuning.

Trained on a modest dataset of 67.5K examples from MSCOCO~\cite{mscoco}, Flickr30K~\cite{flickr30k}, and Conceptual Captions~\cite{conceptual_captions}, DCLIP achieves over 20\% Recall@1 gains in text-to-image retrieval while retaining ~94\% of CLIP’s zero-shot accuracy. It also scales to larger backbones like ViT-L/14, showing retrieval improvements while retaining a meaningful portion of zero-shot generalization—despite the known difficulty of distilling large models.

Our contributions are: 1) a lightweight distillation framework using YOLO-based region features to improve visual grounding without full retraining, 2) an asymmetric student-teacher architecture that preserves CLIP's zero-shot space while improving retrieval performance, and 3) empirical results showing strong retrieval gains with limited data and compute.

\section{Related Work}

\subsection{Contrastive Vision-Language Models}
Our work builds upon contrastive vision-language pretraining, notably CLIP~\cite{clip}, which demonstrated strong zero-shot generalization by learning joint image-text embeddings from web-scale data. Subsequent models like ALIGN~\cite{align}, BLIP~\cite{blip}, and FLORENCE~\cite{florence} further advanced these capabilities by scaling data, refining objectives, or unifying tasks. OpenCLIP~\cite{ilharco2021openclip} facilitated broader research through open-source replications. Despite their success, these models often face limitations in fine-grained understanding and handling variable text lengths. Efforts like Long-CLIP~\cite{zhang2024longclip} (also~\cite{extending_clip_attributes_paper_2503_15485}) and TULIP~\cite{tulip_paper_2410_10034} have addressed text length constraints by modifying positional embeddings or employing specialized adaptation. DCLIP focuses on enhancing fine-grained retrieval by distilling region-informed representations while preserving zero-shot strengths.

\subsection{Region-Aware Vision-Language Understanding}
To capture detailed visual semantics, many approaches incorporate region-level processing, often using object detectors like YOLO~\cite{yolov8} or Faster R-CNN~\cite{ren2015faster}. Early works such as ViLBERT~\cite{vilbert} and LXMERT~\cite{lxmert} used co-attentional transformers over region features and text. Later models like UNITER~\cite{uniter} and VinVL~\cite{vinvl} refined these representations through comprehensive pretraining tasks or improved visual features. Approaches like GLIP~\cite{glip} and RegionCLIP~\cite{regionclip} have further advanced by unifying object detection with phrase grounding or enabling zero-shot region understanding. DCLIP's teacher model leverages YOLOv8~\cite{yolov8} for region identification, but uniquely, this explicit region processing is confined to the teacher; the student model learns to implicitly capture this fine-grained understanding via distillation, maintaining inference efficiency.

\subsection{Knowledge Distillation for Vision-Language Models} 
Knowledge Distillation (KD)~\cite{hinton2015distilling} is a common technique to transfer knowledge from larger teacher models to smaller students. In VLMs, KD has been applied to transfer output distributions, intermediate features~\cite{romero2014fitnets, li2023feature_distillation_vlm}, or relational knowledge~\cite{tian2019crd}. For instance, TinyCLIP~\cite{wu2022tinyclip} distilled both logits and intermediate representations from CLIP. TULIP~\cite{tulip_paper_2410_10034} also employs distillation to adapt its text encoder. DCLIP utilizes a teacher fine-tuned with region-enhanced representations (see Section~\ref{sec:meta_distillation_shortened}) to supervise the student. The student learns these richer semantics through a combination of contrastive learning and direct feature distillation, with an anchor loss for ViT-L variants to preserve zero-shot capabilities.

\subsection{Meta Distillation}\label{sec:meta_distillation_shortened}
Meta-distillation elevates traditional knowledge distillation (KD) by employing a teacher model that is dynamically optimized, adapted, or iteratively improved to provide a superior teaching signal, rather than relying on a static, pre-trained entity. The objective is to enhance the quality and relevance of the knowledge transferred to the student. This paradigm encompasses various strategies for teacher enhancement, such as employing intermediate "teacher assistant" networks~\cite{mirzadeh2020improved}, iterative self-refinement as seen in "Born Again Neural Networks"~\cite{furlanello2018born}, and methods that explicitly optimize the teacher for a specific student or task, like Meta Knowledge Distillation~\cite{yang2022meta_knowledge_distillation} or MetaDistiller~\cite{liu2022metadistiller_flower}. The importance of the teacher's characteristics and training process is also highlighted by research emphasizing teacher consistency~\cite{beyer2022knowledge} and perspectives viewing KD as a teacher-student co-optimization problem~\cite{zhao2022revisiting_kd}. DCLIP's approach aligns with this philosophy: its "meta-teacher" is not an off-the-shelf model but is specifically enhanced through fine-tuning with bidirectional cross-modal attention and YOLO-extracted region features (detailed in Section~\ref{gen_inst}). This targeted enhancement aims to create a teacher that generates more nuanced, fine-grained embeddings, thereby providing a more potent and specialized distillation target.

\section{Methodology and Architecture}
\label{gen_inst}

The DCLIP architecture consists of a teacher-student framework where a frozen cross-modal teacher receives region-level supervision using YOLOv8x~\cite{yolov8}, and a CLIP-based student learns from its embeddings. This distillation enables the student to match the teacher’s fine-grained alignment without ever seeing bounding boxes itself.

\subsection{Cross-Modal Teacher Design}

\begin{figure}[h] 
  \centering
  \includegraphics[width=1\linewidth]{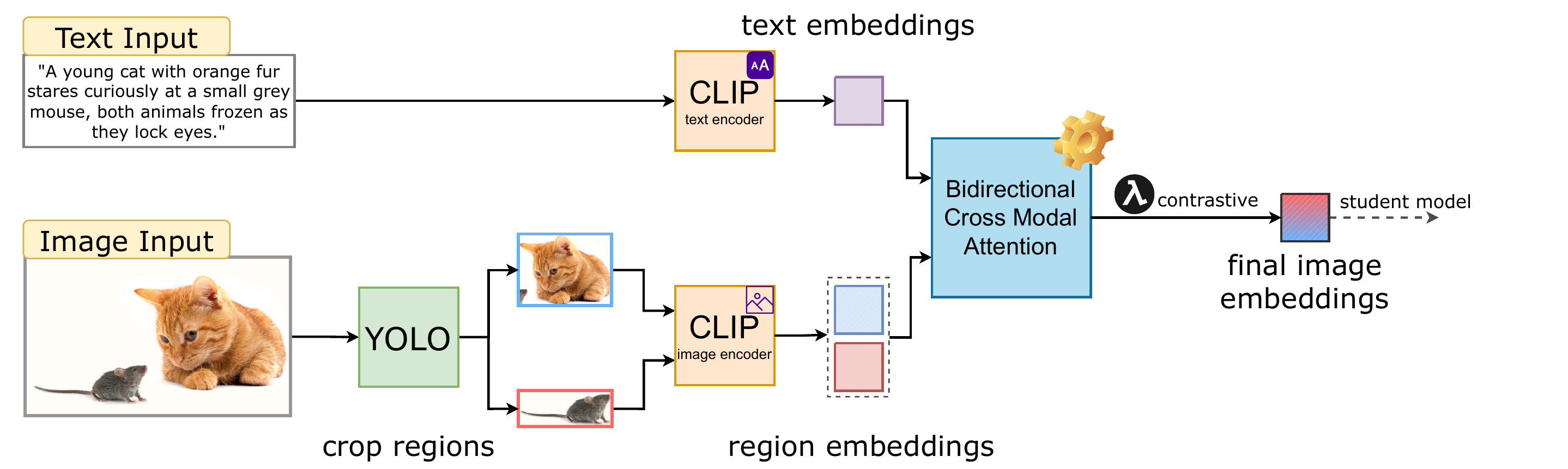}
  \caption{Overview of the DCLIP teacher architecture. Given an image input, YOLO first proposes region-level crops, which are separately encoded by a CLIP image encoder. Simultaneously, the full text is encoded by a CLIP text encoder. Region and text embeddings are fused through a bidirectional cross-modal attention module to produce enriched fine-grained image representations. These final embeddings serve as supervision for student distillation.}
  \label{fig:teacher-architecture}
\end{figure}

The teacher model plays a critical role in distilling detailed image-text alignment signals. It first applies YOLOv8x to extract bounding boxes corresponding to salient image regions. To avoid including irrelevant background or foreground content, each region is weighted using a penalty based on YOLO’s classification confidence, bounding box area, and cosine similarity to the paired text description. These weighted region embeddings are then fused with text tokens using a bidirectional cross-modal transformer, producing semantically rich representations. This additional structure allows the teacher to extract higher-fidelity features than CLIP’s default 224×224 global crop. 

The bidirectional cross-modal attention consists of two separate fine-tuneable multi-head attention layers: one where the text attends to YOLO-derived image regions, and another where the image regions attend to the corresponding text. This symmetrical design ensures that both modalities can learn aligned, context-enriched embeddings, leading to stronger performance across both image-to-text and text-to-image retrieval tasks. 

To create a strong and expressive distillation target, we fine-tune the bidirectional cross-modal attention layers creating our meta-teacher. This allows the teacher to adapt its attention mechanism to the dataset and task, leading to more semantically aligned image-text embeddings. All of the embeddings are aggregated together with a temperature scaled attention to create a global representation. These fine-tuned representations go beyond what static CLIP features offer, encoding cross-modal cues (e.g., object relations, textual grounding) that are essential for high-performance retrieval. The final global embedding $\mathbf{z} \in \mathbb{R}^D$ is computed by temperature-scaled aggregation over the attended patch embeddings $\{\mathbf{h}_i\}_{i=1}^L$:
\[
\mathbf{z} = \sum_{i=1}^{L} \alpha_i \mathbf{h}_i, \quad \text{where} \quad
\alpha_i = \frac{ \exp\left( \frac{ \cos(\mathbf{h}_i, \bar{\mathbf{h}}) }{ \tau } \right) }{ \sum_{j=1}^{L} \exp\left( \frac{ \cos(\mathbf{h}_j, \bar{\mathbf{h}}) }{ \tau } \right) }
\quad \text{and} \quad
\bar{\mathbf{h}} = \frac{1}{L} \sum_{i=1}^{L} \mathbf{h}_i
\]

DCLIP employs a novel meta-teacher distillation framework where a powerful cross-modal teacher generates semantically enriched embeddings with bidirectional attention and region-level supervision. With the finetuned cross modal attention in the meta teacher, we create a semantically aligned distillation target for the student. However, using this teacher directly would require paired image-text inputs at inference time, which compromises CLIP's zero-shot generalization and unimodal usability.

To address this, we distill the teacher's capabilities into a lightweight CLIP student that accepts unimodal inputs. By aligning the student’s representations with those of the teacher via contrastive and cosine-based distillation losses, DCLIP inherits the teacher’s fine-grained alignment benefits while retaining the original CLIP model’s flexibility and zero-shot inference capabilities. This process allows the student to effectively inherit YOLO's supervision without ever seeing bounding boxes itself.

\subsection{Student and Teacher Distillation Losses}


\begin{figure}[h] 
  \centering
  \includegraphics[width=1\linewidth]{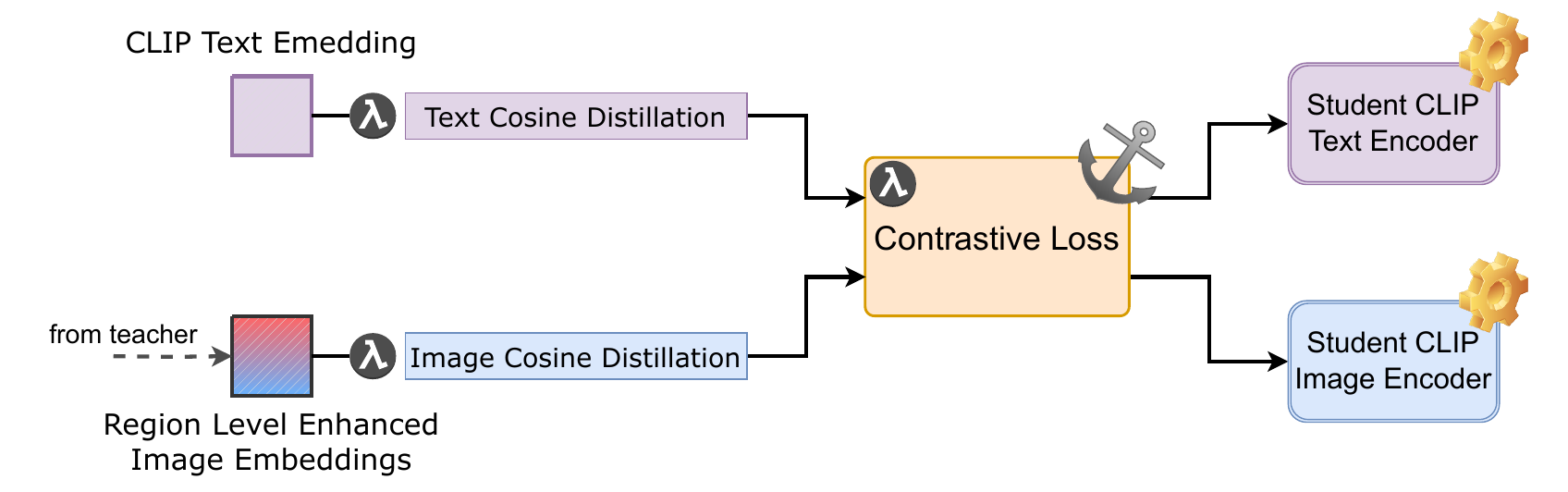}
  \caption{Overview of the DCLIP student architecture. The student never sees the YOLO bounding boxes and only takes in the refined image embeddings from the meta-teacher. Having default CLIP embeddings for the text encoder acts as an anchor to the CLIP space, thus retaining much of the underlying knowledge of CLIP preserving zero shot generalization.}
  \label{fig:teacher-architecture}
\end{figure}

To train both teacher and student, we employ a combination of contrastive and cosine distillation losses. The infoNCE\cite{infoNCE} contrastive loss preserves CLIP’s original discriminative structure by encouraging separation of unrelated image-text pairs, while the cosine losses allow the student to match the teacher’s more detailed embeddings.

\paragraph{Teacher Loss. }The teacher is trained using a standard symmetric InfoNCE~\cite{infoNCE} contrastive loss, which encourages alignment between cross-attended image embeddings and frozen CLIP text embeddings. The loss pulls matched image-text pairs closer while pushing mismatched pairs apart, preserving CLIP’s original semantic structure.

Crucially, DCLIP uses an asymmetric design: only the image encoder is adapted, while the text encoder remains fixed. This prevents the model from shortcutting via trivial alignment and anchors the learned image features to CLIP’s latent space—preserving zero-shot generalization while improving fine-grained retrieval.

\paragraph{Student Loss.}The objective of student training is to construct a retrieval-optimized embedding space that inherits fine-grained alignment from the teacher while maintaining the generalization capacity of the original CLIP model. To achieve this, the student learns from the teacher’s refined embeddings through cosine distillation, aligns semantically matched pairs using a contrastive loss. The combination of these losses ensures that the student can learn task-specific structure (for retrieval) without drifting too far from CLIP’s original general-purpose representations:

\vspace{-0.4cm}
\[
\mathcal{L}_{\text{student}} = 
\mathcal{L}_{\text{contrast}} +
\mathcal{L}_{\text{cos}}^{T} +
\mathcal{L}_{\text{cos}}^{I} 
\]
\vspace{-0.3cm}

Although the InfoNCE contrastive loss formulation is shared with the teacher, the embeddings differ in origin:
\begin{itemize}
    \item The \textbf{teacher} produces $\mathbf{z}^I$,  via a cross-modal transformer with YOLO-guided patch extraction and bidirectional attention. $\mathbf{z}^T$ is a default CLIP text embedding.
    \item The \textbf{student} produces $\mathbf{z}^I$ from its own image encoder finedtuned by the frozen meta teacher's enhanced embeddings, and $\mathbf{z}^T$ is created from CLIP's default text encoder.
\end{itemize}

To distill knowledge from the teacher, we apply cosine distillation losses that directly align student embeddings with their teacher counterparts:

\vspace{-0.2cm} 
\[
\mathcal{L}_{\text{cos}}^{T} = 1 - \text{sim}(\mathbf{z}^{T}_{s}, \mathbf{z}^{T}_{t}), \quad
\mathcal{L}_{\text{cos}}^{I} = 1 - \text{sim}(\mathbf{z}^{I}_{s}, \mathbf{z}^{I}_{t})
\]

Here, the infoNCE contrastive loss plays a pivotal role in preserving the original CLIP structure by ensuring the student does not overfit to the teacher’s embeddings. Without this regularization, the student would risk collapsing to a modality-dependent encoder that sacrifices CLIP’s zero-shot flexibility. The cosine losses guide the student to adopt the teacher’s richer representations, while contrastive alignment maintains discriminative performance and generalization.

\begin{table}[H]
\centering
\small
\setlength{\tabcolsep}{6pt}
\renewcommand{\arraystretch}{1.2} 
\begin{tabular}{p{2cm}p{4cm}p{3.5cm}p{2.5cm}}
\toprule
\textbf{Component} & \textbf{Architecture} & \textbf{Supervision} & \textbf{Inputs} \\
\midrule
\textbf{Teacher} & Cross-modal Transformer with YOLO & Contrastive (InfoNCE) & Image + Text \\
\textbf{Student} & Fine-tuned CLIP encoder & Contrastive (InfoNCE) 
\newline + Cosine Distillation & Image or Text \\
\bottomrule
\end{tabular}
\vspace{0.2cm}
\caption{Summary of DCLIP architecture components. The teacher uses multimodal supervision for fine-grained alignment, while the student learns to approximate its embeddings using unimodal inputs.}
\label{tab:model_summary}
\end{table}

\section{Experimental Set Up}

\textbf{ViT-B.    } We utilize the following mix of data for all ViT-B training and ablations in our experiments. To train the DCLIP model we use a small data set of about 67,500 entries. The mix of datasets are about 50,000 images of MSCOCO 2017, 10,000 images of Flickr30K, and 15,000 images Conceptual Captions entries. This dataset is utilized to train both the student and the teacher. 

\textbf{ViT-L.    } The ViT-L data split is similar to ViT-B but it is just a slight more data. We utilize 125k examples of 80,000 images of MSCOCO, 30,000 images of Flickr30k, and 15,000 images of Conceptual Captions. This increase of data provided less overfitting when distilling with ViT-L.

We observed that larger student models, such as ViT-L/14, are more difficult to distill effectively than smaller models like ViT-B/32 or ViT-B/16. In particular, naive distillation tended to bias the ViT-L embeddings toward text dominance, harming balanced retrieval. To address this, DCLIP modifies the embedding aggregation strategy and type during distillation:

For ViT-B, the lower-dimensional (512-D) feature space allowed effective aggregation around a single semantic cluster. However, for ViT-L (768-D), the increased capacity and higher variance required multi-cluster aggregation. We found that averaging the embeddings from three semantic clusters, rather than one, provided a more robust and balanced supervision signal. Aditionally, the utilization of ROPE\cite{ROPE} embeddings instead of CLIPs default absolute positional embeddings allows for a smoother distillation process and ensures text-to-image retrieval gains while preserving overall image-text retrieval balance. To ensure that ViT-L does not drift away from the original CLIP space, we also add simple cosine distillation preservation loss between the student's image embeddings and the original CLIP ViT-L model to preserve the embedding space.

\section{Results}

\subsection{Evaluation Metrics} 
We evaluate DCLIP across standard retrieval and classification metrics. For evaluation, we utilize the Karpathy\cite{karpathy} testing set for MSCOCO and FLICKR30K, the standard that BLIP and X-VLM uses.  The Karpathy dataset consists of 5,000 images from MSCOCO's validation and training set and a set 1000 images from FLICKR30k. For zero-shot evaluation we use the entire 50,000 images of the Image Net dataset for evaluation. CIFAR-10 and CIFAR-100 datasets are also utilized in evaluation for zero-shot classification to show the robustness of DCLIP.

\paragraph{Recall@K.}
Recall@K measures the proportion of queries for which the correct match appears within the top-K retrieved results. We report Recall@1, Recall@5, and Recall@10 for both Text-to-Image (T→I) and Image-to-Text (I→T) retrieval tasks.

\paragraph{Mean Average Precision (MAP).}
MAP evaluates the overall ranking quality by computing the average precision for each query and averaging over all queries. It rewards models that rank relevant items higher.

\paragraph{Zero-Shot Classification.}
Zero-shot classification assesses the model's ability to generalize to unseen classes without additional training. We report Top-1 and Top-5 accuracy on ImageNet using simple prompts of the form: "This is a photo of a {LABEL}". A prediction is counted as correct if the ground-truth label appears in the top-K predictions.

\subsection{Experimental Results}

\begin{table*}[h]
\centering
\scriptsize
\setlength{\tabcolsep}{5pt}
\begin{tabular}{lcccccccccccc}
\toprule
\multirow{2}{*}{\textbf{Method}} &
\multicolumn{3}{c}{\textbf{Flickr30K (I→T)}} &
\multicolumn{3}{c}{\textbf{Flickr30K (T→I)}} &
\multicolumn{3}{c}{\textbf{MSCOCO (I→T)}} &
\multicolumn{3}{c}{\textbf{MSCOCO (T→I)}} \\
\cmidrule(r){2-4} \cmidrule(r){5-7} \cmidrule(r){8-10} \cmidrule(r){11-13}
& R@1 & R@5 & MAP & R@1 & R@5 & MAP & R@1 & R@5 & MAP & R@1 & R@5 & MAP \\
\midrule
CLIP (ViT-B/32)        & 0.79 & 0.95 & 0.86 & 0.59 & 0.83 & 0.70 & 0.50 & 0.75 & 0.61 & 0.31 & 0.56 & 0.43 \\
DCLIP (ViT-B/32)       & 0.81 & 0.96 & 0.88 & 0.69 & 0.91 & 0.78 & 0.55 & 0.78 & 0.66 & 0.40 & 0.67 & 0.53 \\
\midrule
CLIP (ViT-B/16)        & 0.82 & 0.97 & 0.89 & 0.62 & 0.86 & 0.73 & 0.53 & 0.77 & 0.63 & 0.33 & 0.58 & 0.45 \\
DCLIP (ViT-B/16)       & 0.88 & 0.98 & 0.92 & 0.74 & 0.93 & 0.82 & 0.59 & 0.80 & 0.69 & 0.44 & 0.70 & 0.56 \\
\midrule
CLIP (ViT-L/14)        & 0.85 & 0.97 & 0.90 & 0.65 & 0.87 & 0.75 & 0.56 & 0.79 & 0.67 & 0.36 & 0.61 & 0.48 \\
DCLIP (ViT-L/14)       & 0.87 & 0.98 & 0.91 & 0.75 & 0.94 & 0.83 & 0.59 & 0.81 & 0.69 & 0.47 & 0.72 & 0.59 \\
\midrule
CLIPSelf (ViT-B/16)     & 0.34 & 0.62 &  —   & 0.35 & 0.61 &  —   & 0.19 & 0.39 &  —   & 0.16 & 0.35 &  —   \\
RegionCLIP            & 0.04 & 0.12 &  —   & 0.08 & 0.23 &  —   & 0.02 & 0.07 &  —   & 0.03 & 0.12 &  —   \\
FineCLIP 2.5M (ViT-B/16)& 0.83 & 0.96 &  —   & 0.68 & 0.89 &  —   & 0.55 & 0.79 &  —   & 0.40 & 0.67 &  —   \\
TinyCLIP 8M (ViT-B/16)  & 0.62 &  —   &  —   & 0.42 &  —   &  —   & 0.36 &  —   &  —   & 0.22 &  —   &  —   \\
\bottomrule
\end{tabular}
\vspace{0.2cm}
\caption{Comparison of image-to-text (I→T) and text-to-image (T→I) retrieval on Flickr30K and MSCOCO (Karpathy split) across CLIP, DCLIP, and vision-language baselines. Metrics include Recall@1, Recall@5, and MAP. “—” indicates the metric was not reported in the original paper.}
\label{tab:clip_dclip_comparison_map}
\end{table*}

Because of DCLIP's asymmetric architecture—where only the image encoder is modified while the text encoder remains frozen—our model naturally specializes in the text-to-image retrieval direction. This asymmetry acts as a structural anchor to the original CLIP text space, allowing the student to align more effectively with the unaltered modality. As a result, as shown in Table~\ref{tab:clip_dclip_comparison_map}, we observe consistent and significant gains in text-to-image performance across all CLIP backbones. Although our bidirectional architecture does provide some gains against base CLIP in the image-to-text direction (see Table~\ref{tab:clip_dclip_comparison_map}), showing the effectiveness of the DCLIP architecture of enhancing the model without being the main loss objective.

Among the methods in Table~\ref{tab:clip_dclip_comparison_map}, DCLIP strikes an effective balance between specialized retrieval and broad zero-shot generalization using only 67.5 K image–text pairs. We see substantial text→image R@1 gains of 15–35 pp across ViT-B/32, B/16, and L/14, along with consistent 1–12 pp improvements in image→text retrieval, all without any extra caption generation or large-model supervision. This lightweight distillation makes DCLIP especially practical for low-resource settings. In particular, ViT-B/16 delivers the best overall trade-off, boosting both retrieval directions and retaining about 94 \% of CLIP’s zero-shot classification accuracy.

In contrast to prior small-data distillation approaches that often introduce significant architectural changes or require external supervision, DCLIP’s asymmetric meta-teacher framework scales smoothly even to high-capacity backbones like ViT-L/14. With minimal additional overhead and a straightforward aggregation strategy, it consistently adapts large vision models to specialized retrieval tasks while maintaining strong out-of-domain generalization.


\begin{table*}[h]
\centering
\footnotesize
\resizebox{0.95\linewidth}{!}{
\begin{tabular}{lcccccc}
\toprule
\multirow{2}{*}{\textbf{Model}} & \multicolumn{2}{c}{\textbf{ImageNet-1K}} & \multicolumn{2}{c}{\textbf{CIFAR-10}} & \multicolumn{2}{c}{\textbf{CIFAR-100}} \\
\cmidrule(lr){2-3} \cmidrule(lr){4-5} \cmidrule(lr){6-7}
& Top-1 Acc (\%) & Top-5 Acc (\%) & Top-1 Acc (\%) & Top-5 Acc (\%) & Top-1 Acc (\%) & Top-5 Acc (\%) \\
\midrule
ViT-B/32 (Base CLIP) & 60.08 & 85.03 & 88.21 & 99.35 & 61.50 & 86.31 \\
ViT-B/32 (DCLIP)     & 56.15 & 83.35 & 87.85 & 99.60 & 60.95 & 86.33 \\
\midrule
ViT-B/16 (Base CLIP) & 62.80 & 87.14 & 89.62 & 99.47 & 66.97 & 88.35 \\
ViT-B/16 (DCLIP)     & 59.30 & 85.66 & 89.77 & 99.61 & 65.56 & 88.82 \\
\midrule
ViT-L/14 (Base CLIP) & 69.58 & 89.95 & 94.42 & 99.33 & 75.73 & 92.08 \\
ViT-L/14 (DCLIP)     & 63.79 & 88.14 & 93.92 & 99.53 & 69.90 & 88.02 \\                                                 
\bottomrule
\end{tabular}
}
\vspace{0.2cm}
\caption{
Zero-shot classification accuracy on ImageNet-1K, CIFAR-10, and CIFAR-100 validation sets. DCLIP retains strong zero-shot generalization for ViT-B models, with minimal performance drop across datasets. For ViT-L/14, we use ROPE embeddings to minimize the zero-shot to retrieval trade off and the over powering overfitting of the large embeddings.
}
\label{tab:zero_shot_results}
\end{table*}

From our experiments, we noticed a direct proportionality between the number of epochs trained and the degradation of the underlying CLIP's zero-shot abilities. This reveals a fundamental trade-off in vision-language model specialization: longer teacher training provides clearer supervision signals but drifts further from CLIP's original embedding space.

\begin{figure}[h]
  \centering
  \begin{subfigure}[b]{0.48\textwidth}
    \centering
    \includegraphics[width=\linewidth]{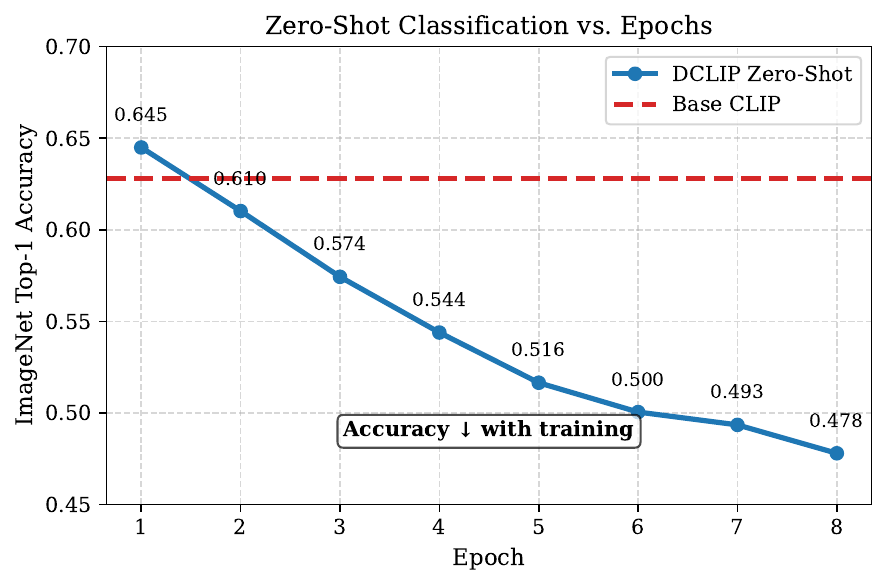}
    \caption{Zero‐Shot Classification vs.\ Epochs}
    \label{fig:zero-shot}
  \end{subfigure}
  \hfill
  \begin{subfigure}[b]{0.48\textwidth}
    \centering
    \includegraphics[width=\linewidth,clip]{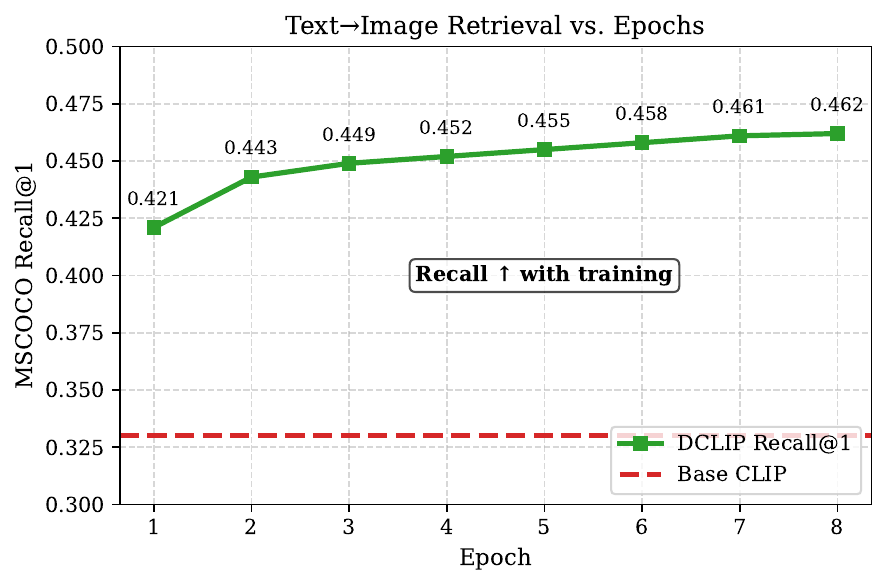}
    \caption{Text→Image Retrieval vs.\ Epochs}
    \label{fig:retrieval}
  \end{subfigure}
  \vspace{0.1cm}  
  \caption{(a) Zero‐shot classification performance over training epochs. (b) Text‐to‐image retrieval improvement over the same schedule. Both graphs recorded on CLIP ViT-B16.}
  \label{fig:main-results}
\end{figure}

Interestingly, this trade-off is not the exact same across retrieval directions. While text to image retrieval consistently improves with longer teacher training, image to text retrieval shows diminishing returns and even slight degradation with overly specialized teachers. This suggests the two pathways respond differently to specialization pressure.

We also observed that validation loss convergence does not directly correlate with optimal retrieval performance. Earlier teacher checkpoints (epochs 2-3) often provide better overall performance when balanced across both retrieval directions and zero-shot preservation, despite higher validation losses.

This finding challenges the conventional wisdom that teacher models should be trained to convergence before distillation. Instead, our experiments suggest a "sweet spot" in teacher training that maximizes retrieval gains while minimizing zero-shot capability degradation.

For additional zero-shot classification behavior on fine-grained domains, we refer readers to the confusion matrices in Appendix E (Fig. 7), which show DCLIP’s strong diagonal structure on CIFAR-10 and more varied but correct predictions on CIFAR-100 under two prompt styles.

\begin{minipage}{0.5\textwidth}
\small
DCLIP maintains strong zero-shot generalization across ImageNet-1K, CIFAR-10, and CIFAR-100 benchmarks. Notably, for ViT-B models, DCLIP achieves comparable or even slightly better Top-5 accuracy than base CLIP on CIFAR datasets (e.g., +0.25\% Top-5 on CIFAR-10 for ViT-B/32). On ImageNet-1K, DCLIP retains approximately 94\% of the original CLIP’s Top-1 zero-shot performance for ViT-B/32 and ViT-B/16. This consistent transferability demonstrates that DCLIP preserves the generalization capacity of CLIP while enriching the visual representations for improved retrieval performance. We further visualize this trade-off in Figure~\ref{fig:pareto-curve}.
\end{minipage}
\hfill
\begin{minipage}{0.45\textwidth}
\centering
\includegraphics[width=\linewidth]{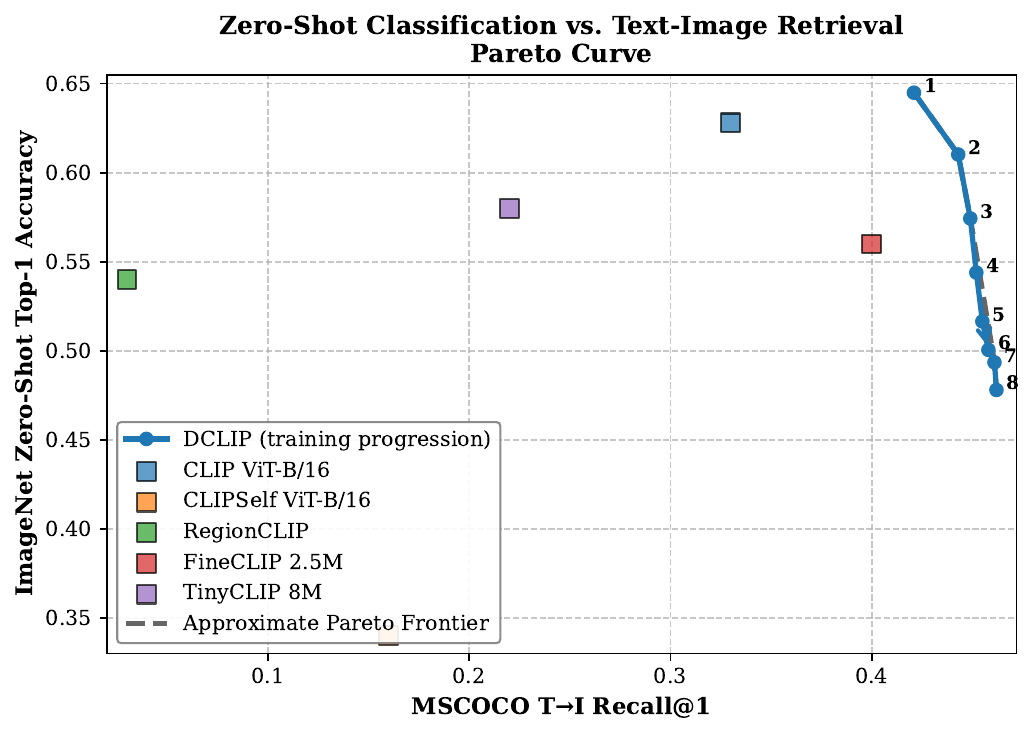}
\captionof{figure}{\scriptsize Pareto curve comparing zero-shot classification vs.\ retrieval performance. DCLIP traces an efficient Pareto frontier under lightweight supervision.}
\label{fig:pareto-curve}
\end{minipage}

ViT-L/14, due to its larger capacity, tends to overfit rapidly to embedding distributions during distillation. That being said, DCLIP manages to still retain 91\% of the original CLIP's zero shot. Without careful control, it can achieve high retrieval performance on specific datasets at the cost of generalization. This behavior partly explains why many prior distillation works avoid using ViT-L: the high computational cost combined with the risk of overfitting makes it less practical. Consequently, ViT-L is typically used when the goal is to maximize retrieval accuracy on a target dataset rather than to maintain broad generalization capabilities.

\section{Ablation Testing and Results}

\begin{table*}[h]
\centering
\scriptsize
\setlength{\tabcolsep}{8pt}
\renewcommand{\arraystretch}{1.0}
\begin{tabular}{lccccccc}
\toprule
\textbf{Ablation Setting} & 
\multicolumn{3}{c}{\textbf{MSCOCO}} & 
\multicolumn{3}{c}{\textbf{Flickr30K}} & 
\textbf{ZS Top-1} \\
\cmidrule(r){2-4} \cmidrule(r){5-7} \cmidrule(r){8-8}
 & T→I R@1 & I→T R@1 & MAP & T→I R@1 & I→T R@1 & MAP & (ImageNet) \\
\midrule
Base CLIP          & 0.33 & 0.53 & 0.45 & 0.61 & 0.81 & 0.72 & 0.63 \\
\midrule
No YOLO (DCLIP)    & 0.36 & 0.54 & 0.48 & 0.65 & 0.83 & 0.75 & \textbf{0.64} \\
No Bidir Attention & 0.41 & 0.56 & 0.53 & 0.68 & 0.84 & 0.78 & 0.61 \\
No Teacher    & 0.40 & 0.54 & 0.52 & 0.69 & 0.81 & 0.79 & 0.58 \\
No CMA             & 0.41 & 0.56 & 0.53 & 0.68 & 0.83 & 0.89 & 0.59 \\
\midrule
Full DCLIP         & \textbf{0.44} & \textbf{0.59} & \textbf{0.56} & \textbf{0.74} & \textbf{0.88} & \textbf{0.82} & 0.59 \\
\bottomrule
\end{tabular}
\vspace{0.2cm}
\caption{Ablation results on MSCOCO and Flickr30K using ViT-B/16 (Karpathy split). Metrics are Recall@1 (T→I and I→T) and MAP(T→I); “ZS Top-1” is zero-shot ImageNet accuracy. Removing components like YOLO, bidirectional attention (CMA), or the meta-teacher impacts retrieval performance and generalization.}
\label{tab:ablation_results}
\end{table*}

We perform critical ablation results on DCLIP with CLIP's ViT-B-16 model. Our overall best performance and distilled model from our experimental results. Our ablations show how each component works together to help DCLIP create strong embeddings for retrieval.

\paragraph{No Region-Level Embeddings.}
Region-level embeddings prove to be a significant improvement over CLIP’s original center-crop strategy. By leveraging YOLO to extract semantically rich regions, the model gains access to more informative image cues, particularly benefiting image-to-text retrieval. Interestingly, removing YOLO slightly improves zero-shot classification performance. This suggests that without region supervision, the model remains closer to the original CLIP embedding space, enhancing generalization. This behavior aligns with our findings with a  standard fine tuning setup, indicating that minimal architectural shifts favor zero-shot retention.

\paragraph{No Bidirectional Cross-Attention.}
Cross-attention plays a key role in aligning region-level visual features with text representations. In this ablation, we replace bidirectional attention with a simple average of patch and text embeddings. Importantly, we preserve the original temperature-scaled attention for global representation. The results show that while region embeddings alone outperform CLIP’s vanilla features, the absence of cross-attention leads to suboptimal alignment. This highlights that interaction between modalities is critical—not just the richness of features, but how well they are integrated.

\paragraph{No Meta-Teacher Guidance.}
In ViT-B/16, distillation from a well-trained teacher significantly boosts performance. Removing the teacher leads to weaker performance, even when region embeddings and cross-attention are retained. This indicates that smaller models benefit from strong external guidance to shape their embedding space effectively. Conversely, our findings with ViT-L/14 reveal that such guidance can become a liability: large models tend to overfit quickly to their teacher, limiting generalization. Thus, distillation is most effective in compact models, where learning capacity and inductive biases are better complemented by external supervision.

While ViT-B/16 presents the best trade-off under limited supervision, our experiments with ViT-L/14 revealed a surprising and important behavior. Contrary to expectations, training ViT-L/14 without a distillation teacher yielded significantly more balanced retrieval performance, especially in image-to-text tasks and zero-shot preservation. When trained with a teacher, the model rapidly overfit — leading to sharp gains in one retrieval direction but disproportionate degradation elsewhere.

\begin{minipage}[t]{0.62\textwidth}
\vspace{0pt}
\textbf{ViT-L/14 (No Teacher) Observations.} Distilling larger models like ViT-L/14 poses unique challenges due to their tendency to overfit rapidly during fine-tuning. Our experiments show that without teacher guidance, ViT-L distillation improves text-to-image retrieval moderately, but at the cost of degradation in image-to-text retrieval and overall embedding balance.
\end{minipage}
\hfill
\begin{minipage}[t]{0.35\textwidth}
\vspace{0pt}
\centering
\scriptsize
\setlength{\tabcolsep}{3pt}
\renewcommand{\arraystretch}{1.1}
\begin{tabular}{lcccc}
\toprule
\textbf{Dataset} & \textbf{Dir} & \textbf{R@1} & \textbf{R@5} & \textbf{MAP} \\
\midrule
\multirow{2}{*}{MSCOCO}
  & T$\rightarrow$I & 0.45 & 0.71 & 0.57 \\
  & I$\rightarrow$T & 0.58 & 0.81 & 0.67 \\
\midrule
\multirow{2}{*}{Flickr30K}
  & T$\rightarrow$I & 0.73 & 0.93 & 0.82 \\
  & I$\rightarrow$T & 0.85 & 0.98 & 0.90 \\
\midrule
\multicolumn{2}{l}{\textbf{INet-ZS}} & 54.10 & 79.54 & -- \\
\bottomrule
\end{tabular}
\vspace{0.1cm}
\captionof{table}{\scriptsize ViT-L/14 ROPE (No Teacher) results (Karpathy split). All values rounded to two decimals.}
\end{minipage}

These results suggest that while larger ViT models inherently learn stronger region-level embeddings, they require structured supervision—such as from a cross-modal teacher—to avoid overfitting and maintain generalization. DCLIP’s architecture principles, particularly multi-cluster aggregation and teacher-guided contrastive alignment, thus scale effectively even to high-capacity vision-language models.

\section{Limitations}
While DCLIP achieves significant improvements in retrieval under limited supervision, it relies on external region proposals generated by pretrained object detectors such as YOLO during training. This guided supervision introduces a dependency that may limit flexibility in settings where object annotations or reliable detectors are unavailable. Furthermore, as YOLO detectors are typically optimized for in-domain datasets (e.g., MSCOCO), their generalization to out-of-domain distributions remains limited. Applying DCLIP to significantly different domains would require either hand-annotating new datasets with YOLO or adapting alternative region proposal methods, which can be time-consuming and labor-intensive. Additionally, while DCLIP demonstrates strong retention of zero-shot generalization and retrieval gains even with high-capacity models like ViT-L/14, distilling very large vision models remains an open challenge due to their greater sensitivity to overfitting at the region level.  
Additionally, due to compute constraints, DCLIP was evaluated primarily on relatively small-scale datasets (MSCOCO, Flickr30k, Conceptual Captions 67.5k). The scalability of DCLIP to larger web-scale datasets, such as LAION-400M or CC12M, remains an open question for future investigation.

\section{Conclusion}

In this paper, we introduced \textbf{DCLIP}, a distilled image-text model that leverages cross-modal supervision, adaptive region proposals, and tailored loss functions to enhance retrieval performance while preserving zero-shot classification capabilities. Through extensive experiments, we showed that DCLIP outperforms CLIP and other distilled baselines on downstream retrieval tasks across MS-COCO and Flickr30K, particularly under compute- and memory-constrained settings. Our approach highlights the effectiveness of integrating spatial grounding into student models and reveals promising avenues for efficient multimodal representation learning. Overall, DCLIP demonstrates that careful distillation of spatial and semantic knowledge from CLIP can yield lightweight yet competitive alternatives for real-world deployment.

\begin{ack}
Use unnumbered first level headings for the acknowledgments. All acknowledgments
go at the end of the paper before the list of references. Moreover, you are required to declare
funding (financial activities supporting the submitted work) and competing interests (related financial activities outside the submitted work).
More information about this disclosure can be found at: \url{https://neurips.cc/Conferences/2025/PaperInformation/FundingDisclosure}.

Do {\bf not} include this section in the anonymized submission, only in the final paper. You can use the \texttt{ack} environment provided in the style file to automatically hide this section in the anonymized submission.
\end{ack}

{
\small


\newpage 
\appendix

\section*{Appendix: DCLIP Further Details} 

\renewcommand{\thesubsection}{\Alph{subsection}}
\renewcommand{\thesubsubsection}{\Alph{subsection}.\arabic{subsubsection}} 

This appendix provides supplementary information to the main paper, including detailed hyperparameter settings, model architecture specifics, dataset information, loss function formulations, evaluation metric definitions, additional experimental results.

\subsection{Implementation Details}
\label{app:implementation_details_lettered}

\paragraph{Training Hyperparameters}
\label{app:hyperparameters_lettered}
This section outlines the specific hyperparameters used for training the teacher and student models. The Adam optimizer was used for all training, with a batch size of 32.

\begin{table}[h]
\centering
\small
\resizebox{0.90\linewidth}{!}{%
\begin{tabular}{lcc}
\toprule
\textbf{Parameter} & \textbf{ViT-B} & \textbf{ViT-L} \\
\midrule
Teacher Training Epochs & 5 & 1 \\
Student Training Epochs & 2 & 2 \\
Teacher Learning Rate & $1 \times 10^{-5}$ & $1 \times 10^{-5}$ \\
Student Learning Rate & $1 \times 10^{-6}$ & $1 \times 10^{-6}$ \\
Attention Heads & 8 & 12 \\
Optimizer & Adam & Adam \\
Global Patch Aggregation Method & Mean (1 cluster) & Mean (3 clusters) \\
Amount of Training Data & 67.5k & 120k \\
Teacher Loss & Contrastive & Contrastive \\
Student Loss & Contrastive + Cosine  & Contrastive + Cosine + ZS Anchor Loss \\
Batch Size & 32 & 32 \\
Hidden Dimension & 512 & 768 \\
\bottomrule
\end{tabular}
} 
\vspace{0.2em}
\parbox{0.90\linewidth}{
  \caption{Comparison of training hyperparameters and architectural choices between ViT-B and ViT-L during DCLIP distillation. Aggregation method refers to how region embeddings are averaged during supervision. ZS stands for Zero Shot.}
  \label{tab:vitb_vitl_training_hyperparameters}
}
\end{table}

For ViT-L/14 student distillation, the anchor loss refers to the "simple cosine distillation preservation loss between the student's image embeddings and the original CLIP ViT-L model's image embeddings." The main paper also mentions that for ViT-L, "averaging the embeddings from three semantic clusters, rather than one, provided a more robust and balanced supervision signal."

\paragraph{Model Architecture Details}
\label{app:architecture_filled_lettered}
The DCLIP framework consists of a teacher-student setup. The teacher model utilizes YOLO-extracted image regions and a cross-modal attention mechanism, while the student is a fine-tuned CLIP image encoder.

\paragraph{Cross-Modal Teacher Design.}
The teacher model (visualized in Figure \ref{fig:teacher_architecture_appendix}) first applies YOLOv8x to extract bounding boxes. Each region is weighted using a penalty based on YOLO’s classification confidence, bounding box area, and cosine similarity to the paired text description. These weighted region embeddings are fused with text tokens using a bidirectional cross-modal transformer. This transformer consists of two separate fine-tuneable multi-head attention layers: one where text attends to image regions, and another where image regions attend to text. This allows the teacher to adapt its attention mechanism, leading to more semantically aligned image-text embeddings. Specifics like the number of transformer layers ($N_L$), attention heads ($N_H$), and hidden dimension ($D_H$) for this module were illustratively set to $N_L=2, N_H=8, D_H=512$ in previous discussions, and should be confirmed from implementation.

\begin{figure}[H] 
  \centering
  \includegraphics[width=0.9\linewidth]{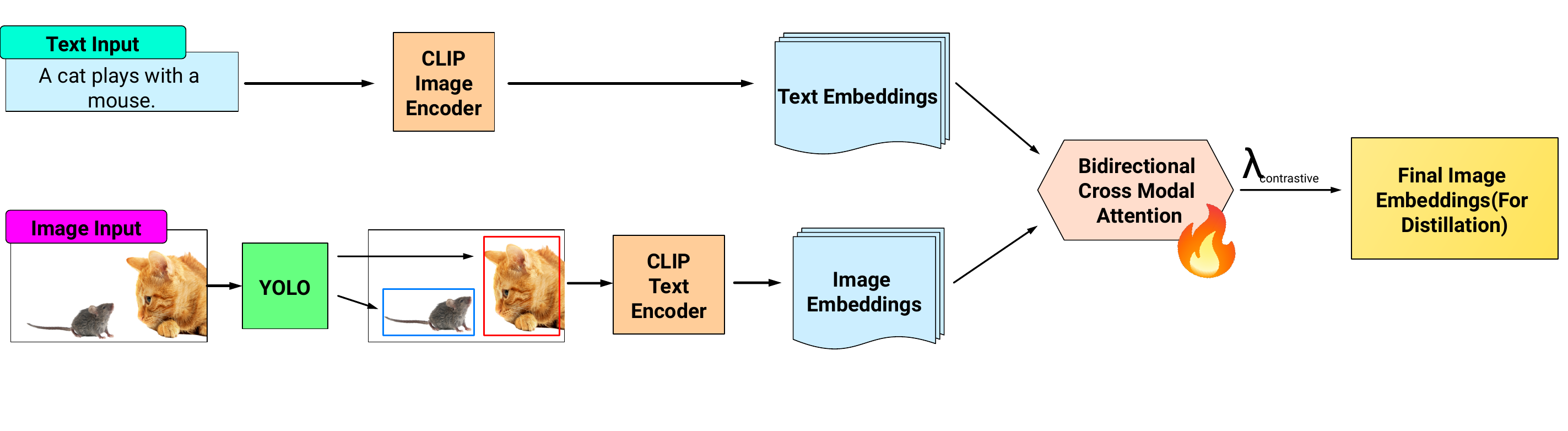} 
  \captionsetup{width=0.9\linewidth} 
  \caption{DCLIP Teacher Architecture Overview. YOLO extracts region-level features which are fused with text embeddings through bidirectional cross-modal attention to produce fine-grained aligned image representations. (Diagram based on Figure 1 from provided document)}
  \label{fig:teacher_architecture_appendix}
\end{figure}

\paragraph{Student Model.}
The student model (visualized in Figure \ref{fig:student_architecture_appendix}) is a standard CLIP image encoder (ViT-B/32, ViT-B/16, or ViT-L/14) whose image encoder weights are fine-tuned. The text encoder remains frozen and identical to the original CLIP text encoder. The student learns from the teacher's refined embeddings and aims to inherit YOLO's supervision without directly processing bounding boxes.

\begin{figure}[H] 
  \centering
  \includegraphics[width=0.9\linewidth]{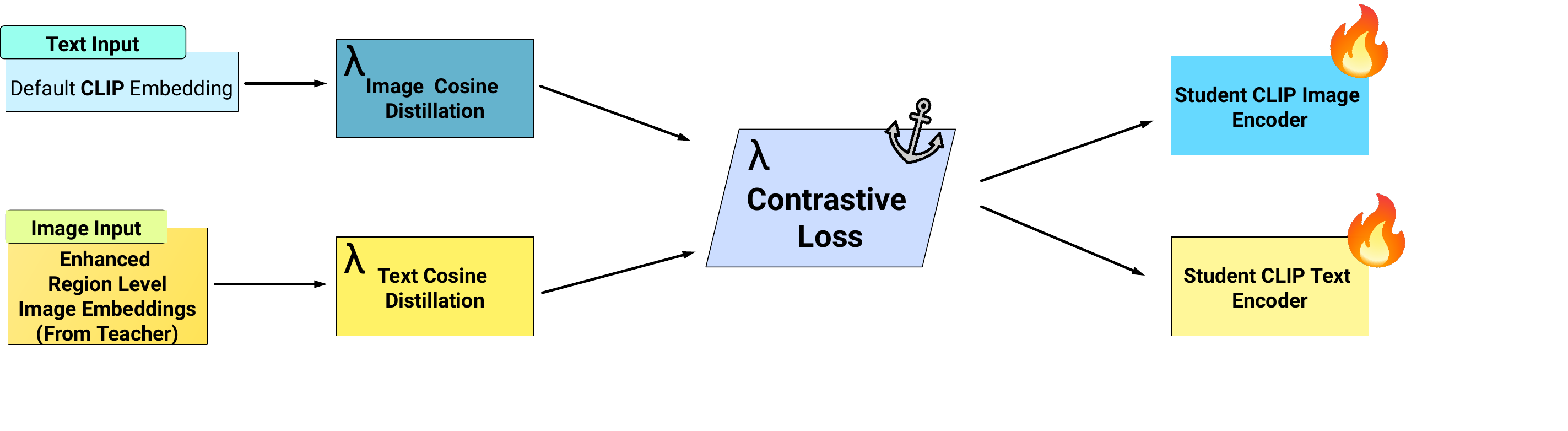} 
  \captionsetup{width=0.9\linewidth} 
  \caption{DCLIP Student Architecture Overview. The student never sees the YOLO bounding boxes and only takes in the refined image embeddings from the meta-teacher. Having default CLIP embeddings for the text encoder acts as an anchor to the CLIP space, thus retaining a lot of the underlying knowlegde of CLIP preserving zero shot generalization. (Diagram based on Figure 2 from provided document)}
  \label{fig:student_architecture_appendix}
\end{figure}

\paragraph{Computational Efficiency}  To demonstrate DCLIP’s practicality on modest hardware, we cache both YOLO region proposals and CLIP embeddings, drastically reducing redundant computation. All ViT-B experiments (both B-32 and B-16) run end-to-end—including fine-tuning of the student and teacher—on a single NVIDIA RTX 2070 Super (8 GB) in roughly 1.5 hours per epoch. Even the larger ViT-L/14 models train in about 2 hours per epoch on an NVIDIA RTX 5090 (32 GB). These results show that DCLIP’s meta-teacher distillation can be conducted within reasonable GPU-hour and memory budgets, making it accessible to researchers with consumer-grade hardware.

\paragraph{Software and Libraries}
\label{app:software_filled_lettered}
The DCLIP framework was implemented using Python. Key libraries typically include:
\begin{itemize}
    \item PyTorch
    \item Transformers (Hugging Face) (for CLIP models)
    \item Ultralytics YOLO (for YOLOv8x)
    \item NumPy
\end{itemize}

\subsection{Dataset Details}
\label{app:dataset_details_filled_lettered}

\paragraph{Data Curation and Splits}
\label{app:data_curation_filled_lettered}
The training data for DCLIP was curated from publicly available datasets, totaling approximately 67,500 entries for the primary experiments (Section 4 of provided document):
\begin{itemize}
    \item MSCOCO 2017: ~42,000 image-text pairs.
    \item Flickr30K: ~10,000 image-text pairs.
    \item Conceptual Captions: ~15,000 image-text pairs.
\end{itemize}

The training data for ViT-L to prevent less over specialization was about 120,000 entries for the primary experiments:
\begin{itemize}
    \item MSCOCO 2017: ~85,000 image-text pairs.
    \item Flickr30K: ~20,000 image-text pairs.
    \item Conceptual Captions: ~15,000 image-text pairs.
\end{itemize}
This dataset was used for both teacher and student training. Evaluation was performed on the Karpathy test splits for MSCOCO and Flickr30K. Zero-shot evaluation used ImageNet, CIFAR-10, and CIFAR-100 (as per original main paper's Section 5.1, confirm if still accurate).

\paragraph{Preprocessing}
\label{app:preprocessing_filled_lettered}
All patches were resized to $224 \times 224$ pixels and normalized using the mean and standard deviation specific to the CLIP pretrained models. For the teacher, regions extracted by YOLO are cropped and resized.
Text captions were tokenized using the default CLIP tokenizer. The maximum sequence length was capped at 77 tokens.

\subsection{Loss Function Formulations}
\label{app:loss_details_lettered}

\paragraph{Teacher and Student InfoNCE Loss}
\label{app:teacher_loss_lettered}
The teacher is trained using a symmetric InfoNCE contrastive loss (Equation \ref{eq:contrastive_appendix}, from Section 3.2 of provided document). Given a batch of $N$ cross-attended image embeddings $\{ \mathbf{z}^I_i \}_{i=1}^N$ (derived from YOLO regions and cross-modal attention) and default CLIP text embeddings $\{ \mathbf{z}^T_j \}_{j=1}^N$, all embeddings are L2-normalized ($\hat{\mathbf{z}}^I_i, \hat{\mathbf{z}}^T_j$). The similarity matrix $\mathbf{L} \in \mathbb{R}^{N \times N}$ is:
\[
\mathbf{L}_{i,j} = \frac{\hat{\mathbf{z}}^I_i \cdot \hat{\mathbf{z}}^T_j}{\tau}
\]
where $\tau$ is a temperature scaling parameter. The loss is:
\begin{equation}
\mathcal{L}_{\text{contrast}}^{\text{teacher}} = \frac{1}{2N} \sum_{i=1}^{N} \text{CE}(\mathbf{L}_{i,:}, i) +
\frac{1}{2N} \sum_{j=1}^{N} \text{CE}(\mathbf{L}_{:,j}, j) \label{eq:contrastive_appendix}
\end{equation}
where $\text{CE}(\cdot, \cdot)$ is the cross-entropy loss. This encourages the teacher to maintain global semantic structure by aligning matched pairs and separating mismatched ones.

\paragraph{Student Loss}
\label{app:student_loss_lettered}
The student loss function (from Section 3.2 of provided document) is:
\[
\mathcal{L}_{\text{student}} = 
\mathcal{L}_{\text{contrast}}^{\text{student}} +
\mathcal{L}_{\text{cos}}^{T} +
\mathcal{L}_{\text{cos}}^{I} 
\]
$\mathcal{L}_{\text{contrast}}^{\text{student}}$ is an InfoNCE loss similar to the teacher's, but using the student's fine-tuned CLIP image encoder embeddings ($\mathbf{z}_{s}^{I}$) and frozen CLIP text embeddings ($\mathbf{z}_{s}^{T}$).
The cosine similarity distillation losses are:
\[
\mathcal{L}_{\text{cos}}^{T} = 1 - \text{sim}(\mathbf{z}^{T}_{s}, \mathbf{z}^{T}_{t})
\]
\[
\mathcal{L}_{\text{cos}}^{I} = 1 - \text{sim}(\mathbf{z}^{I}_{s}, \mathbf{z}^{I}_{t})
\]
Here, $\mathbf{z}^{T}_{s}$ are student text embeddings (from frozen CLIP text encoder), $\mathbf{z}^{T}_{t}$ are teacher text embeddings (identical to $\mathbf{z}^{T}_{s}$). $\mathbf{z}^{I}_{s}$ are student image embeddings, and $\mathbf{z}^{I}_{t}$ are teacher image embeddings (from the cross-modal teacher). The InfoNCE loss regularizes the student, preventing overfitting to teacher embeddings and preserving zero-shot capabilities. The asymmetric distillation (only image encoder adapted) anchors the loss to the original language prior. For ViT-L/14, an additional anchor loss was also used.

\subsection{Evaluation Metric Definitions}
\label{app:eval_metrics_lettered}
(From Section 5.1 of provided document)

\paragraph{Recall@K}
\label{app:recall_k_lettered}
Recall@K measures the proportion of queries for which the correct match appears in the top-K retrieved results.
\[
\text{Recall@}K = \frac{1}{N} \sum_{i=1}^{N} \text{Success@K}(i)
\]
where $N$ is the number of queries, and $\text{Success@K}(i) = 1$ if the correct item for query $i$ is among the top-K retrieved results, and 0 otherwise. Reported for $K \in \{1, 5, 10\}$.

\paragraph{Mean Average Precision (MAP)}
\label{app:map_lettered}
MAP evaluates the ranking quality across all queries.
\[
\text{MAP} = \frac{1}{N} \sum_{i=1}^{N} \text{AP}(i)
\]
where Average Precision $\text{AP}(i)$ for query $i$ is:
\[
\text{AP}(i) = \frac{1}{|\mathcal{R}_i|} \sum_{k=1}^{K_{max\_rank}} P_i(k) \times \delta(k)
\]
Here, $\mathcal{R}_i$ is the set of relevant items for query $i$, $P_i(k)$ is the precision at cutoff $k$, $\delta(k) = 1$ if the item at rank $k$ is relevant, and $K_{max\_rank}$ is the total number of items ranked.

\paragraph{Zero-Shot Classification Accuracy (Top-K)}
\label{app:zs_accuracy_lettered}
Zero-shot performance evaluates generalization to unseen classes using prompts like "This is a photo of a \{LABEL\}".

\[
\text{Top-K Accuracy} = \frac{1}{N} \sum_{i=1}^{N} \text{Correct@}K(i)
\]

where $\text{Correct@K}(i) = 1$ if the ground truth label $y_i$ is among the top-K predictions for image $x_i$, and 0 otherwise. Reported for Top-1 and Top-5.

\subsection{Additional Experimental Results and Observations}
\label{app:additional_results_filled_lettered}

\begin{figure}[H]
  \centering
  \begin{subfigure}[t]{0.48\linewidth}
    \centering
    \includegraphics[width=\linewidth]{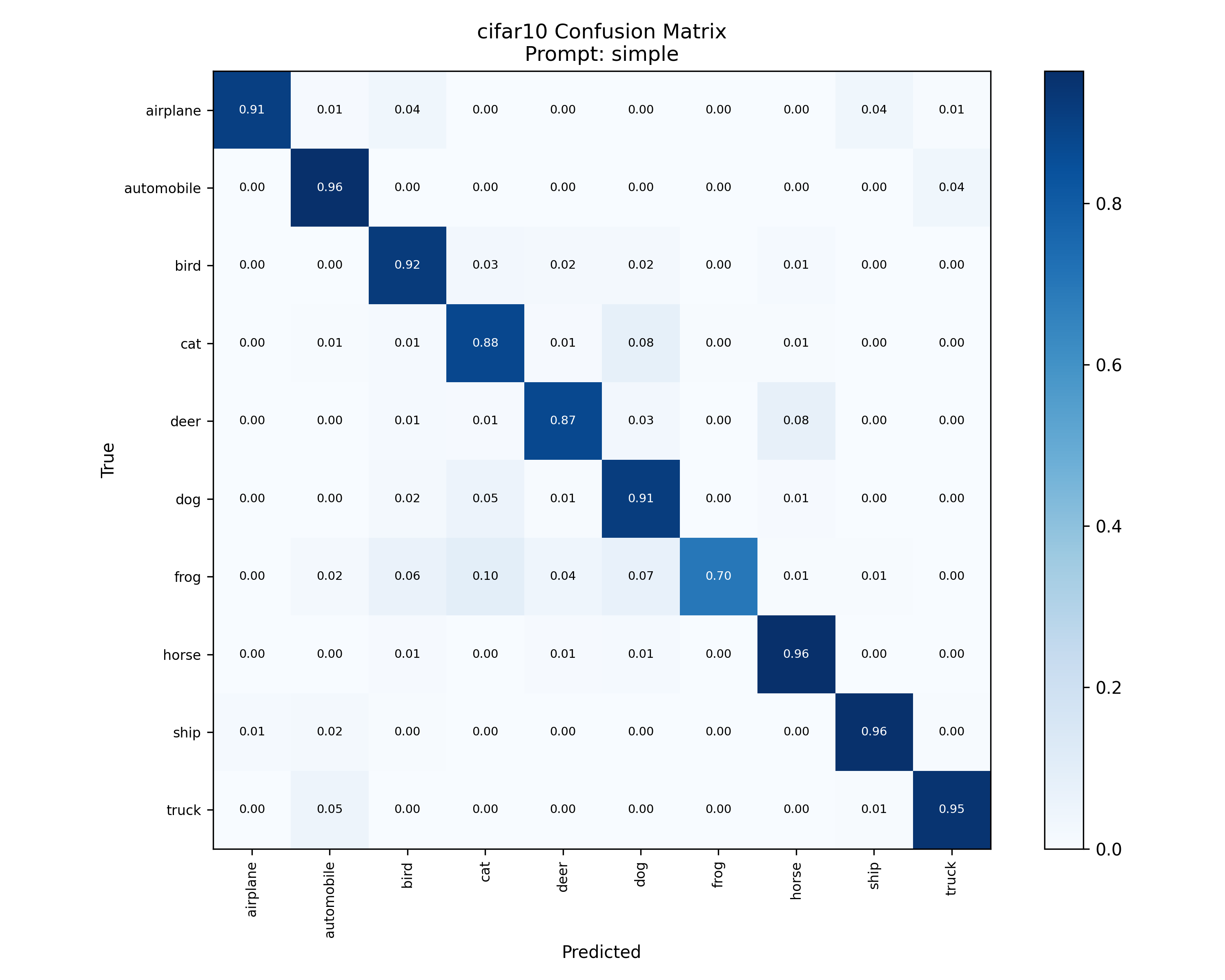}
    \caption{CIFAR-10 (prompt: “simple”).}
    \label{fig:cifar10-confusion}
  \end{subfigure}\hfill
  \begin{subfigure}[t]{0.48\linewidth}
    \centering
    \includegraphics[width=\linewidth]{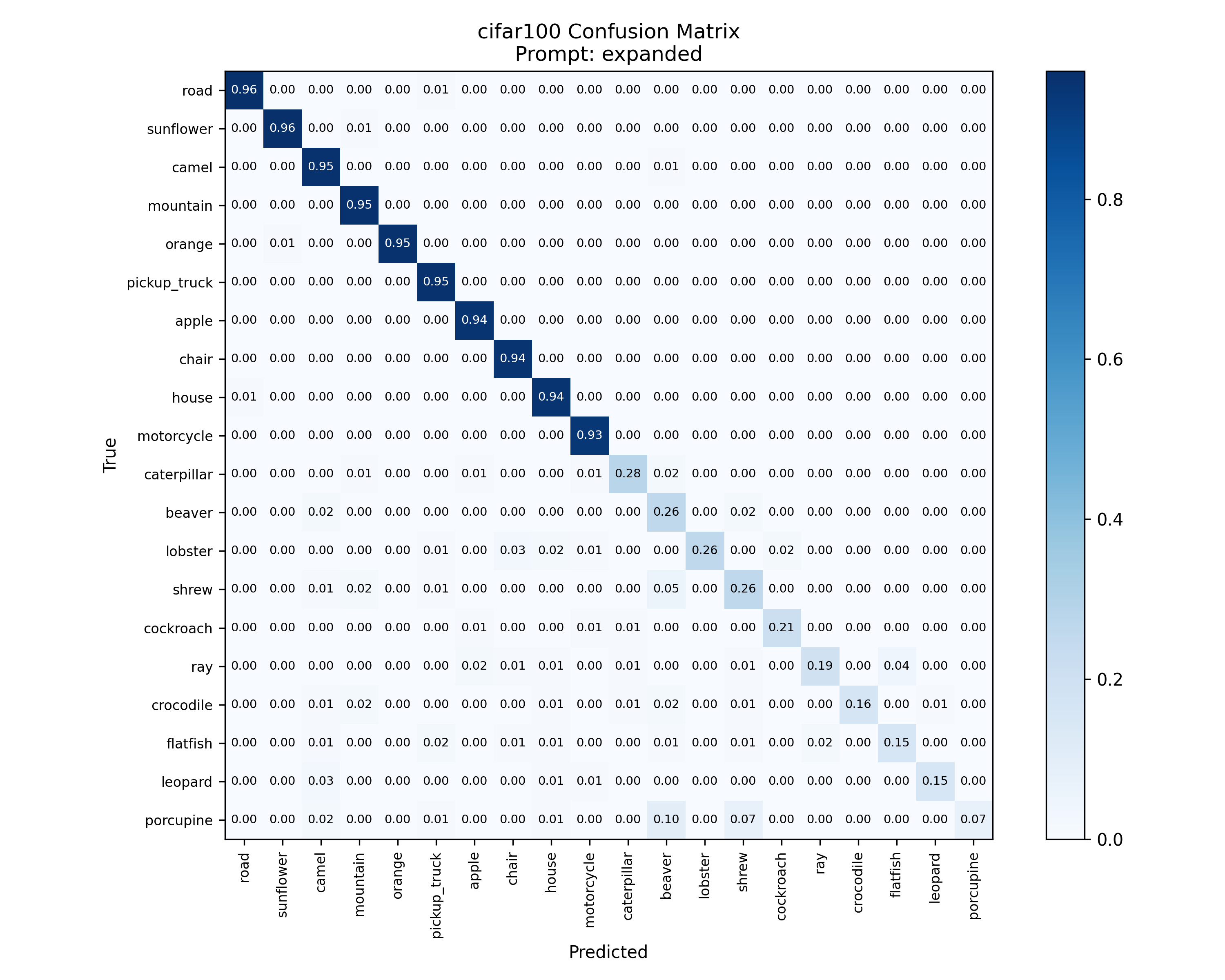}
    \caption{CIFAR-100 (prompt: “expanded”).}
    \label{fig:cifar100-confusion}
  \end{subfigure}
  \captionsetup{font=small}
  \parbox{0.90\linewidth}{
  \caption[Zero-shot CIFAR confusion matrices]{Zero-shot confusion matrices for DCLIP ViT-B/16 on CIFAR-10 and CIFAR-100 under two prompt styles.}
  \label{fig:cifar-confusions}
  }
\end{figure}

\paragraph{Open Vocabulary Prompts}
For the CIFAR-10 confusion matrix we utilized a simple prompt that simply predicted the class of the image, "a photo of a {PREDICTION}". For CIFAR-100's confusion matrix we used the expanded prompt that added a little bit more context, "a photo of a {} type or organism".

\paragraph{Zero-Shot Classification Confusion Matrices.} Figure~\ref{fig:cifar-confusions} presents confusion matrices for zero-shot classification on CIFAR-10 and CIFAR-100 using DCLIP ViT-B/16. On CIFAR-10, the model exhibits strong diagonal dominance, indicating high class separability, especially for structured categories like "automobile," "dog," and "ship." Minor confusion occurs between visually similar classes (e.g., "cat" and "dog," "deer" and "horse"). On CIFAR-100, performance is more varied due to the larger number of fine-grained classes, but DCLIP still correctly identifies distinct categories such as "road," "sunflower," and "motorcycle." These results further support DCLIP’s ability to generalize in fine-grained classification tasks, even without task-specific supervision.

\paragraph{Training Observations}
(From discussion in Section 5.2 of provided document)
Experiments indicated a trade-off: longer teacher training (more epochs) provided clearer supervision signals but could lead to the teacher drifting further from CLIP's original embedding space, potentially degrading the student's zero-shot capabilities. Text-to-image retrieval consistently improved with longer teacher training, while image-to-text retrieval showed diminishing returns or slight degradation with overly specialized teachers. Validation loss convergence did not directly correlate with optimal retrieval performance; earlier teacher checkpoints (e.g., epochs 2-3) sometimes offered a better balance across retrieval directions and zero-shot preservation. This suggests a "sweet spot" in teacher training rather than training to full convergence.

\begin{table}[H]
\centering
\parbox{0.90\linewidth}{
\caption{Approximate parameter counts for CLIP Vision Transformers (Image Encoder only).}
}
\label{tab:model_params_appendix_filled_lettered}
\small
\begin{tabular}{lc}
\toprule
\textbf{Model Backbone} & \textbf{Parameters (Image Encoder, Approx.)} \\
\midrule
ViT-B/32 & $\sim$86 M \\
ViT-B/16 & $\sim$86 M \\
ViT-L/14 & $\sim$304 M \\
\bottomrule
\end{tabular}
\end{table}
The cross-modal attention module in the teacher adds parameters but is not part of the final student model used for inference.

\paragraph{ITM Loss DCLIP} 
In previous experiments we utilized a image text matching(ITM) loss in both the student and the teacher. The ITM loss, utilized by BLIP, creates a binary classification head that is trained based on hard negatives that allows the classifier to make decisions on if it thinks that caption belongs with that specific image. We performed this experiment by distilling with ViT-B-32 and training on the previous mentioned 67.5k dataset. Evaluation was not done on the Karpathy split but rather 1000 images of Flickr30k and MSCOCO 2014.

\newpage

\begin{table}[h]
\centering
\scriptsize
\setlength{\tabcolsep}{8pt}
\renewcommand{\arraystretch}{1.1}
\begin{tabular}{lccc}
\toprule
\textbf{Metric} & \textbf{CLIP} & \textbf{DCLIP} & \textbf{Gain} \\
\midrule
\multicolumn{4}{c}{\textbf{MSCOCO (1k Images from COCO 2014)}} \\
\midrule
T→I R@1    & 0.50 & \textbf{0.61} & +22.9\% \\
T→I R@5    & 0.80 & \textbf{0.88} & +10.6\% \\
T→I MAP    & 0.63 & \textbf{0.73} & +15.9\% \\
I→T R@1    & 0.70 & \textbf{0.74} & +5.0\%  \\
I→T R@5    & 0.91 & \textbf{0.92} & +1.1\%  \\
I→T MAP    & 0.79 & \textbf{0.82} & +3.8\%  \\
\midrule
\multicolumn{4}{c}{\textbf{Flickr30K (1k Subset)}} \\
\midrule
T→I R@1    & 0.59 & \textbf{0.70} & +18.9\% \\
T→I R@5    & 0.85 & \textbf{0.92} & +8.2\%  \\
T→I MAP    & 0.70 & \textbf{0.80} & +14.3\% \\
I→T R@1    & 0.77 & \textbf{0.83} & +7.9\%  \\
I→T R@5    & 0.96 & \textbf{0.96} & +0.0\%  \\
I→T MAP    & 0.85 & \textbf{0.89} & +4.7\%  \\
\midrule
\multicolumn{4}{c}{\textbf{ImageNet Zero-Shot Classification}} \\
\midrule
Top-1 Acc. & 0.60 & \textbf{0.53} & 88\% retention \\
Top-5 Acc. & 0.85 & \textbf{0.81} & 95\% retention \\
\bottomrule
\end{tabular}
\vspace{0.2cm}
\parbox{0.90\linewidth}{
\caption{Retrieval and zero-shot classification performance for CLIP vs.\ DCLIP on Karpathy-split benchmarks. Gains are relative to the CLIP baseline; “retention” indicates the percentage of original zero-shot accuracy preserved.}
}
\label{tab:results_flat}
\end{table}

\paragraph{Analysis of ITM Loss} We experimented with adding the BLIP‐style ITM head to both teacher and student in the ViT-B/32 DCLIP setting, trained on the 67.5 K sample subset and evaluated on 1 K splits. While ITM yielded a +1.2 pp gain in Text→Image R@1 (0.61→0.62), it also accelerated zero-shot Top-1 decay on ImageNet by 3 pp after 5 epochs and increased training time per epoch by 15 

\paragraph{DCLIP ViT-B-16 on 220k Dataset}

\begin{table}[htbp]
\centering
\scriptsize
\setlength{\tabcolsep}{8pt} 
\renewcommand{\arraystretch}{1.1} 
\begin{tabular}{@{}l c c c@{}}
\toprule
\textbf{Metric} & \textbf{Base CLIP} & \textbf{Custom Model (DCLIP)} & \textbf{Gain} \\
\midrule
\multicolumn{4}{c}{\textbf{MSCOCO}} \\
\midrule
T$\to$I R@1    & 0.33 & 0.41 & +24.4\% \\
T$\to$I R@5    & 0.58 & 0.67 & +14.6\% \\
T$\to$I R@10   & 0.69 & 0.77 & +11.3\% \\
T$\to$I MAP    & 0.45 & 0.53 & +17.5\% \\
\cmidrule(lr){1-4}
I$\to$T R@1    & 0.53 & 0.53 & +1.1\%  \\
I$\to$T R@5    & 0.77 & 0.77 & +0.3\%  \\
I$\to$T R@10   & 0.85 & 0.85 & +0.8\%  \\
I$\to$T MAP    & 0.63 & 0.64 & +0.8\%  \\
\midrule
\multicolumn{4}{c}{\textbf{Flickr30K}} \\
\midrule
T$\to$I R@1    & 0.62 & 0.68 & +9.6\%  \\
T$\to$I R@5    & 0.86 & 0.90 & +5.4\%  \\
T$\to$I R@10   & 0.92 & 0.95 & +3.4\%  \\
T$\to$I MAP    & 0.73 & 0.78 & +7.4\%  \\
\cmidrule(lr){1-4}
I$\to$T R@1    & 0.82 & 0.82 & –0.6\%  \\
I$\to$T R@5    & 0.97 & 0.96 & –0.4\%  \\
I$\to$T R@10   & 0.99 & 0.99 & –0.3\%  \\
I$\to$T MAP    & 0.89 & 0.88 & –0.4\%  \\
\midrule
\multicolumn{4}{c}{\textbf{ImageNet Zero-Shot Classification}} \\
\midrule
Top-1 Acc.     & 0.60 & 0.57 & 94\% retention \\
Top-5 Acc.     & 0.85 & 0.85 & 100\% retention \\
\bottomrule
\end{tabular}
\vspace{0.2cm}
\parbox{0.90\linewidth}{
\caption{Retrieval and zero-shot classification performance for Base CLIP vs.\ Custom Model (DCLIP) on MSCOCO, Flickr30K, and ImageNet. Gains are relative to Base CLIP; “retention” indicates the percentage of original zero-shot accuracy preserved.}
\label{tab:retrieval_performance_final}
}
\end{table}

To set up this experiment we extended the dataset to about 220k images. The split was ~100k MSCOCO, ~30k Flickr, ~50k Conceptual Captions, and ~50k Visual Genome\cite{krishna2017visualgenome}. This extended dataset convers a broad area of domains in computer vision.

Table~\ref{tab:retrieval_performance_final} shows when we increase the distillation dataset from 67.5 K to 200 K image–text pairs, DCLIP’s retrieval performance on the held-out Karpathy splits improves modestly (e.g. +3–5 pp R@1 across MSCOCO and Flickr30K). This behavior suggests that with more (and potentially noisier) distillation examples, the student becomes more specialized to the retrieval objectives at the expense of the broad, out-of-domain generalization that CLIP’s original zero-shot head provides. In other words, the extra data helps DCLIP better model fine-grained image–text alignment—pushing up recall metrics—but also causes it to “overfit” those retrieval signals and drift away from the teacher’s class-agnostic representations.



\newpage

\end{document}